\documentclass[sigconf]{acmart}

\AtBeginDocument{
  \providecommand\BibTeX{{
    \normalfont B\kern-0.5em{\scshape i\kern-0.25em b}\kern-0.8em\TeX}}}

\usepackage{textcomp}
\usepackage{xcolor}
\usepackage[labelformat=simple]{subcaption}

\usepackage{graphicx}
\usepackage[utf8]{inputenc}
\usepackage{listings}
\usepackage{algorithm}
\usepackage{algorithmicx,algpseudocode}
\usepackage{xspace}
\usepackage{shortcuts}
\usepackage{graphicx}
\usepackage{multirow}
\usepackage{makecell}
\usepackage[export]{adjustbox}
\usepackage{caption}
\usepackage{color}
\usepackage{hyperref}
\usepackage{booktabs}

\usepackage{enumitem}

\def\BibTeX{{\rm B\kern-.05em{\sc i\kern-.025em b}\kern-.08em
    T\kern-.1667em\lower.7ex\hbox{E}\kern-.125emX}}

\definecolor{mygreen}{rgb}{0,0.6,0}
\definecolor{mygray}{rgb}{0.5,0.5,0.5}
\definecolor{mymauve}{rgb}{0.58,0,0.82}

\renewcommand\footnotetextcopyrightpermission[1]{}
\begin{document}
\fancyhead{}

\title{\codename: Learning an Assembly Language Model for Instruction Embedding}

 \pagestyle{plain}

\author{Xuezixiang Li}
\affiliation{
  \institution{University of California Riverside}
  \country{Riverside, CA 92521, USA}
  }
\email{xli287@ucr.edu}

\author{Yu Qu}
\affiliation{
  \institution{University of California Riverside}
  \country{Riverside, CA 92521, USA}
  }
\email{yuq@ucr.edu}

\author{Heng Yin}
\affiliation{
  \institution{University of California Riverside}
  \country{Riverside, CA 92521, USA}
  }
\email{heng@cs.ucr.edu}

\begin{abstract}

Deep learning has demonstrated its strengths in numerous binary analysis tasks, including function boundary detection, binary code search, function prototype inference, value set analysis, etc. When applying deep learning to binary analysis tasks, we need to decide what input should be fed into the neural network model. More specifically, we need to answer how to represent an instruction in a fixed-length vector. The idea of automatically learning instruction representations is intriguing, but the existing schemes fail to capture the unique characteristics of disassembly. These schemes ignore the complex intra-instruction structures and mainly rely on control flow in which the contextual information is noisy and can be influenced by compiler optimizations.

In this paper, we propose to pre-train an assembly language model called \codename for generating general-purpose instruction embeddings by conducting self-supervised training on large-scale unlabeled binary corpora. \codename utilizes three pre-training tasks to capture various characteristics of assembly language. These training tasks overcome the problems in existing schemes, thus can help to generate high-quality representations. We conduct both intrinsic and extrinsic evaluations, and compare \codename with other instruction embedding schemes. \codename has the best performance for intrinsic metrics, and outperforms the other instruction embedding schemes for all downstream tasks.

\end{abstract}

\ccsdesc[500]{Security and privacy~Software reverse engineering}
\ccsdesc[300]{Theory of computation~Program analysis}
\ccsdesc[300]{Computing methodologies~Knowledge representation and reasoning}


\keywords{Deep Learning, Binary Analysis, Language Model, Representation Learning}
\maketitle
\section{Introduction}

Recently, we have witnessed a surge of research efforts that leverage deep learning to tackle various binary analysis tasks, including function boundary identification~\cite{Shin:sec15}, binary code similarity detection~\cite{Gemini:ccs17,zuo2018InnerEYE,alphaDiff:ase18,yu2020order,pei2020trex}, function prototype inference~\cite{Chua17EKLAVYA}, value set analysis~\cite{DeepVSA}, malware classification~\cite{malconv}, etc. Deep learning has shown noticeably better performances over the traditional program analysis and machine learning methods.

When applying deep learning to these binary analysis tasks, the first design choice that should be made is: what kind of input should be fed into the neural network model? Generally speaking, there are three choices: we can either directly feed raw bytes into a neural network (e.g., the work by Shin et al.~\cite{Shin:sec15}, $\alpha$Diff~\cite{alphaDiff:ase18}, DeepVSA~\cite{DeepVSA}, and MalConv~\cite{malconv}), or feed manually-designed features (e.g., Gemini~\cite{Gemini:ccs17} and Instruction2Vec~\cite{Lee2017ins2vec}), or automatically learn to generate a vector representation for each instruction using some representation learning models such as word2vec (e.g., InnerEye~\cite{zuo2018InnerEYE} and EKLAVYA~\cite{Chua17EKLAVYA}), and then feed the representations (embeddings) into the downstream models. 

Compared to the first two choices, automatically learning\linebreak instruction-level representation is more attractive for two reasons: (1) it avoids manually designing efforts, which require expert knowledge and may be tedious and error-prone; and (2) it can learn higher-level features rather than pure syntactic features and thus provide better support for downstream tasks. To learn instruction-level representations, researchers adopt algorithms (e.g., word2vec~\cite{word2vec2013} and PV-DM~\cite{le2014distributed}) from Natural Language Processing (NLP) domain, by treating binary assembly code as natural language documents.

Although recent progress in instruction representation learning (instruction embedding) is encouraging, there are still some unsolved problems which may greatly influence the quality of instruction embeddings and limit the quality of downstream models:

First, existing approaches ignore the complex internal formats of instructions. For instance, in x86 assembly code, the number of operands can vary from zero to three; an operand could be a CPU register, an expression for a memory location, an immediate constant, or a string symbol; some instructions even have implicit operands, etc. Existing approaches either ignore this structural information by treating an entire instruction as a word (e.g., InnerEye~\cite{zuo2018InnerEYE} and EKLAVYA~\cite{Chua17EKLAVYA}) or only consider a simple instruction format (e.g., Asm2Vec~\cite{dingasm2vec}). Second, existing approaches use Control Flow Graph (CFG) to capture contextual information between instructions (e.g., Asm2Vec~\cite{dingasm2vec}, InnerEye~\cite{zuo2018InnerEYE}, and the work by Yu et al.~\cite{yu2020order}). However, the contextual information on control flow can be noisy due to compiler optimizations, and cannot reflect the actual dependency relations between instructions.

Moreover, in recent years, pre-trained deep learning models~\cite{qiu2020pre} are increasingly attracting attentions in different fields such as Computer Vision (CV) and Natural Language Processing (NLP). The intuition of pre-training is that with the development of deep learning, the numbers of model parameters are increasing rapidly. A much larger dataset is needed to fully train model parameters and to prevent overfitting. Thus, pre-trained models (PTMs) using \textit{large-scale unlabeled corpora} and \textit{self-supervised training} tasks have become very popular in some fields such as NLP. Representative deep pre-trained language models in NLP include BERT~\cite{devlin2019bert}, GPT~\cite{radford2018improving}, RoBERTa~\cite{liu2019roberta}, ALBERT~\cite{lan2019albert}, etc. Considering the naturalness of programming languages~\cite{hindle2012naturalness,allamanis2018survey} including assembly language, it has great potential to pre-train an assembly language model for different binary analysis tasks.

To solve the existing problems in instruction representation learning and capture the underlying characteristics of instructions, in this paper, we propose a pre-trained assembly language model called \codename\footnote{\codename stands for \textbf{P}re-trained \textbf{A}ssembly \textbf{L}anguage \textbf{M}odel for Ins\textbf{TR}uction \textbf{E}mb\textbf{E}dding} for general-purpose instruction representation learning. \codename is based on the BERT~\cite{devlin2019bert} model but pre-trained with newly designed training tasks exploiting the inherent characteristics of assembly language.

We are not the first to utilize the BERT model in binary analysis. For instance, Yu et al.~\cite{yu2020order} proposed to take CFG as input and use BERT to pre-train the token embeddings and block embeddings for the purpose of binary code similarity detection. Trex~\cite{pei2020trex} uses one of BERT's pre-training tasks -- Masked Language Model (MLM) to learn program execution semantics from functions' micro-traces (a form of under-constrained dynamic traces) for binary code similarity detection.

Contrast to the existing approaches, our goal is to develop a pre-trained assembly language model for \textit{general-purpose} instruction representation learning. Instead of only using MLM on control flow, \codename uses three training tasks to exploit special characteristics of assembly language such as instruction reordering introduced by compiler optimizations and long range data dependencies. The three training tasks work at different granularity levels to effectively train \codename to capture internal formats, contextual control flow dependency, and data flow dependency of instructions.

Experimental results show that \codename can provide high quality general-purpose instruction embeddings. Downstream applications can directly use the generated embeddings in their models. A static embedding lookup table can be generated in advance for common instructions. Such a pre-trained, general-purpose language model scheme is especially useful when computing resources are limited such as on a lower-end or embedded devices.

We design a set of intrinsic and extrinsic evaluations to systematically evaluate \codename and other instruction embedding models. In intrinsic evaluations, we conduct outlier detection and basic block similarity search. In extrinsic evaluations, we use several downstream binary analysis tasks, which are binary code similarity detection, function type signatures analysis, and value set analysis, to evaluate \codename and the baseline models. Experimental results show that \codename has the best performance in intrinsic evaluations compared with the existing models. In extrinsic evaluations, \codename outperforms the other instruction embedding models and also significantly improves the quality of the downstream applications. We conclude that \codename can effectively generate high-quality instruction embedding which is helpful for different downstream binary analysis tasks.

In summary, we have made the following contributions:
\begin{itemize}

\item We lay out several challenges in the existing schemes in instruction representation learning.

\item We pre-train an assembly language model called \codename to generate general-purpose instruction embeddings and overcome the existing challenges.

\item We propose to use three pre-training tasks for \codename embodying the characteristics of assembly language such as reordering and long range data dependency.

\item We conduct extensive empirical evaluations and demonstrate that \codename outperforms the other instruction embedding models and also significantly improves the accuracy of downstream binary analysis tasks.

\item We plan to release the source code of \codename, the pre-trained model, and the evaluation framework to facilitate the follow-up research in this area.

\end{itemize}

To facilitate further research, we have made the source code and pre-trained \codename model publicly available at \url{https://github.com/palmtreemodel/PalmTree}.

\section{Background} \label{section:Background}

In this section, we firstly summarize existing approaches and background knowledge of instruction embedding. Then we discuss some unsolved problems of the existing approaches. Based on the discussions, we summarize representative techniques in this field.

\subsection{Existing Approaches}

Based on the embedding generation process, existing approaches can be classified into three categories: raw-byte encoding, manually-designed encoding, and learning-based encoding.

\subsubsection{Raw-byte Encoding}

The most basic approach is to apply a simple encoding on the raw bytes of each instruction, and then feed the encoded instructions into a deep neural network. One such encoding is ``one-hot encoding'', which converts each byte into a 256-dimensional vector. One of these dimensions is 1 and the others are all 0. MalConv~\cite{malconv} and DeepVSA~\cite{DeepVSA} take this approach to classify malware and perform coarse-grained value set analysis, respectively. 

One instruction may be several bytes long. To strengthen the sense of an instruction, DeepVSA further concatenates the one-hot vectors of all the bytes belonging to an instruction, and forms a vector for that instruction. 

Shin et al.~\cite{Shin:sec15} take a slightly different approach to detect function boundaries. Instead of a one-hot vector, they encode each byte as a 8-dimensional vector, in which each dimension presents a corresponding digit in the binary representation of that byte. For instance, the \textbf{\texttt{0x90}} will be encoded as 
$$ [\ 1\ 0\ 0\ 1\ 0\ 0\ 0\ 0\ ]$$

In general, this kind of approach is simple and efficient, because it does not require disassembly, which can be computationally expensive. Its downside, however, is that it does not provide any semantic level information about each instruction. For instance, we do not even know what kind of instruction it is, and what operands it operates on. While the deep neural networks can probably learn some of this information by itself, it seems very difficult for the deep neural networks to completely understand all the instructions.

\subsubsection{Manual Encoding of Disassembled Instructions} \label{section: Manually Selected Encoding}

Knowing that disassembly carries more semantic information about an instruction, this approach first disassembles each instruction and encodes some features from the disassembly. 

Li et al.~\cite{li2019graph} proposed a very simple method, which only extracts opcode to represent an instruction, and encodes each opcode as a one-hot vector. Unfortunately, this method completely ignores the information from operands.
Instruction2Vec~\cite{Lee2017ins2vec} makes use of both opcode and operand information. Registers, addresses, and offsets are encoded in different ways, and then concatenated to form a vector representation. Each instruction is encoded as a nine-dimensional feature vector. An instruction is divided into tokens, and tokens are encoded as unique index numbers. While an opcode takes one token, a memory operand takes up to four tokens, including base register, index register, scale, and displacement. 

\ignore{

\begin{figure}[ht]
\centering
\includegraphics[width=0.7\linewidth]{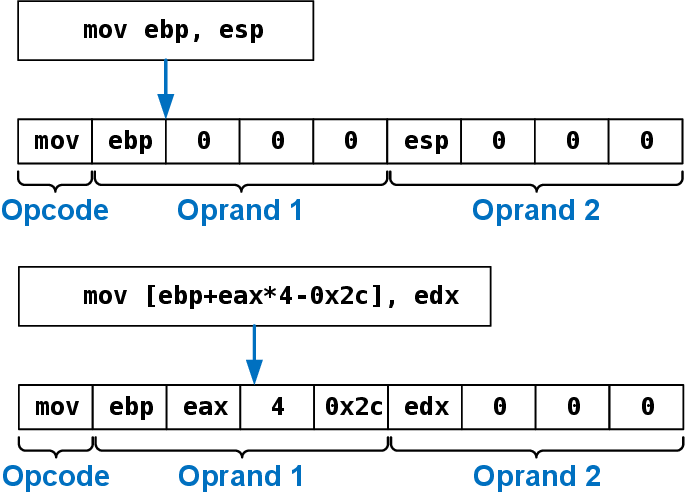}
\caption{Examples of Instruction2Vec~\cite{Lee2017ins2vec}}
\label{figure:instruction2vec}
\end{figure}
}

While this approach is able to reveal more information about opcode and operands for each instruction than raw-byte encoding, it does not carry higher-level semantic information about each instruction. For instance, it treats each opcode instruction equally unique, without knowing that \textbf{\texttt{add}} and \textbf{\texttt{sub}} are both arithmetic operations thus they are more similar to each other than \textbf{\texttt{call}}, which is a control transfer operation. Although it is possible to manually encode some of the higher-level semantic information about each instruction, it requires tremendous expert knowledge, and it is hard to get it right.

\subsubsection{Learning-based Encoding}

Inspired by representation learning in other domains such as NLP (e.g., word2vec~\cite{word2vec2013,mikolov2013efficient}), we would like to automatically learn a representation for each instruction that carries higher-level semantic information. Then this instruction-level representation can be used for any downstream binary analysis tasks, achieving high analysis accuracy and generality. 

Several attempts have been made to leverage word2vec~\cite{word2vec2013} to automatically learn instruction-level representations (or embeddings), for code similarity detection~\cite{zuo2018InnerEYE,Massarelli2019Safe} and function type inference~\cite{Chua17EKLAVYA}, respectively. The basic idea of this approach is to treat each instruction as a word, and each function as a document. By applying a word2vec algorithm (Skip-gram or CBOW~\cite{mikolov2013efficient, word2vec2013}) on the disassembly code in this way, we can learn a continuous numeric vector for each instruction. 

In order to detect similar functions in binary code, Asm2Vec~\cite{dingasm2vec} makes use of the PV-DM model~\cite{le2014distributed} to generate instruction embeddings and an embedding for the function containing these instructions simultaneously. Unlike the above approach that treats each instruction as a word, Asm2Vec treats each instruction as one opcode and up to two operands and learns embeddings for opcodes and operands separately.

\subsection{Challenges in Learning-based Encoding}
\label{subsection:challenges}

While the learning-based encoding approach seems intriguing, there exist several challenges.

\ignore{For these challenges, Reviewer 2 pointed out: the authors outline three main challenges that should be addressed to provide a more insightful and useful assembly instruction embedding. However, such needs are not always well-motivated. For instance, the authors state that related work does not exploit the internal format of instructions and, therefore, treats instructions with multiple or individual operands (with different types) as a word or just a simple instruction format. While I do see the benefit in distinguishing types (e.g., register vs memory operands), I believe the authors should stress more what one would miss should this information not be captured properly. Similar reasoning holds for the other two challenges - the authors seem to provide a subjective opinion rather than an objective treatment of the limitations of such embeddings. This is particularly relevant in the introduction, which should set the scene more convincingly and in an objective manner.}

\ignore{Thus we should improve the discussion (better with more measurement results and statistics). We should also improve the introduction part accordingly.}

\subsubsection{Complex and Diverse Instruction Formats} 

Instructions (especially those in CISC architectures) are often in a variety of formats, with additional complexities. \autoref{lst:case1} gives several examples of instructions in x86. 

\begin{center}
\begin{minipage}{0.95\linewidth}
\begin{lstlisting}[label=lst:case1,basicstyle=\scriptsize\ttfamily\bfseries,caption=Instructions are complex and diverse, language={[x86masm]Assembler},escapechar=!, frame=single]
 ; memory operand with complex expression
 mov [ebp+eax*4-0x2c], edx
 ; three explicit operands, eflags as implicit operand
 imul [edx], ebx, 100 
 ; prefix, two implicit memory operands 
 rep movsb                    
 ; eflags as implicit input
 jne 0x403a98
\end{lstlisting}
\end{minipage}
\end{center}

In x86, an instruction can have between 0 to 3 operands. An operand can be a CPU register, an expression for a memory location, an immediate constant, or a string symbol. A memory operand is calculated by an expression of ``\textbf{\texttt{base}}$+$\textbf{\texttt{index}}$\times$\textbf{\texttt{scale}}$+$\textbf{\texttt{displacement}}''. While \textbf{\texttt{base}} and \textbf{\texttt{index}} are CPU registers, \textbf{\texttt{scale}} is a small constant number and \textbf{\texttt{displacement}} can be either a constant number or a string symbol. All these fields are optional. As a result, memory expressions vary a lot. Some instructions have implicit operands. Arithmetic instructions change \textbf{\texttt{EFLAGS}} implicitly, and conditional jump instructions take \textbf{\texttt{EFLAGS}} as an implicit input.

A good instruction-level representation must understand these internal details about each instruction. Unfortunately, the existing learning-based encoding schemes do not cope with these complexities very well. Word2vec, adopted by some previous efforts~\cite{zuo2018InnerEYE,Massarelli2019Safe,Chua17EKLAVYA}, treats an entire instruction as one single word, totally ignoring these internal details about each instruction. 

Asm2Vec~\cite{dingasm2vec} looks into instructions to a very limited degree. It considers an instruction having one opcode and up to two operands. In other words, each instruction has up to three tokens, one for opcodes, and up to two for operands. A memory operand with an expression will be treated as one token, and thus it does not understand how a memory address is calculated. It does not take into account other complexities, such as prefix, a third operand, implicit operands, \texttt{\textbf{EFLAGS}}, etc.

\begin{center}
\begin{minipage}{0.95\linewidth}
\begin{lstlisting}[label=lst:case2,basicstyle=\scriptsize\ttfamily\bfseries,caption=Instructions can be reordered, language={[x86masm]Assembler},escapechar=!, frame=single]
; prepare the third argument for function call          
mov rdx, rbx
; prepare the second argument for function call             
mov rsi, rbp
; prepare the first argument for function call            
mov rdi, rax
; call memcpy() function         
call memcpy
; test rbx register (this instruction is reordered)             
test rbx, rbx
; store the return value of memcpy() into rcx register
mov rcx, rax
; conditional jump based on EFLAGS from test instruction
je  0x40adf0
\end{lstlisting}
\end{minipage}
\end{center}

\subsubsection{Noisy Instruction Context}\label{subsubsec:context}

The context is defined as a small number of instructions before and after the target instruction on the control-flow graph. These instructions within the context often have certain relations with the target instruction, and thus can help infer the target instruction's semantics.

\begin{table*}[t]
\centering
\small
\caption{Summary of Approaches}
\label{table: Approaches Overview}
\begin{tabular}{llccc}
\toprule
\textbf{Name} & \textbf{Encoding} & \textbf{Internal Structure} & \textbf{Context} & \textbf{Disassembly Required}\\ 
\midrule
DeepVSA~\cite{DeepVSA} & 1-hot encoding on raw-bytes & no & no & no \\
Instruction2Vec~\cite{Lee2017ins2vec}& manually designed & yes & no & yes \\
InnerEye~\cite{zuo2018InnerEYE} & word2vec & no & control flow & yes \\
Asm2Vec~\cite{dingasm2vec} & PV-DM & partial & control flow & yes \\
\codename (this work) & BERT & yes & control flow \& data flow & yes \\
\bottomrule
\end{tabular}
\end{table*}

While this assumption might hold in general, compiler optimizations tend to break this assumption to maximize instruction level parallelism. In particular, compiler optimizations (e.g., ``-fschedule-insns'', ``-fmodulo-sched'', ``-fdelayed-branch'' in GCC) seek to avoid stalls in the instruction execution pipeline by moving the load from a CPU register or a memory location further away from its last store, and inserting irrelevant instructions in between.

\autoref{lst:case2} gives an example. The \textbf{\texttt{test}} instruction at Line 10 has no relation with its surrounding \textbf{\texttt{call}} and \textbf{\texttt{mov}} instructions. The \textbf{\texttt{test}} instruction, which will store its results into \textbf{\texttt{EFLAGS}}, is moved before the \textbf{\texttt{mov}} instruction by the compiler, such that it is further away from the \textbf{\texttt{je}} instruction at Line 14, which will use (load) the \textbf{\texttt{EFLAGS}} computed by the \textbf{\texttt{test}} instruction at Line 10. From this example, we can see that contextual relations on the control flow can be noisy due to compiler optimizations.

Note that instructions also depend on each other via data flow (e.g., lines 8 and 12 in~\autoref{lst:case2}). Existing approaches only work on control flow and ignore this important information. On the other hand, it is worth noting that most existing PTMs cannot deal with the sequence longer than 512 tokens~\cite{qiu2020pre} (PTMs that can process longer sequences, such as Transformer XL~\cite{dai2019transformer}, will require more GPU memory), as a result, even if we directly train these PTMs on instruction sequences with MLM, it is hard for them capture long range data dependencies which may happen among different basic blocks. Thus a new pre-training task capturing data flow dependency is desirable.

\subsection{Summary of Existing Approaches}

\autoref{table: Approaches Overview} summarizes and compares the existing approaches, with respect to which encoding scheme or algorithm is used, whether disassembly is required, whether instruction internal structure is considered, and what context is considered for learning. In summary, raw-byte encoding and manually-designed encoding approaches are too rigid and unable to convery higher-level semantic information about instructions, whereas the existing learning-based encoding approaches cannot address challenges in instruction internal structures and noisy control flow. 

\begin{figure*}[t]
\centering
\includegraphics[width=0.8\linewidth]{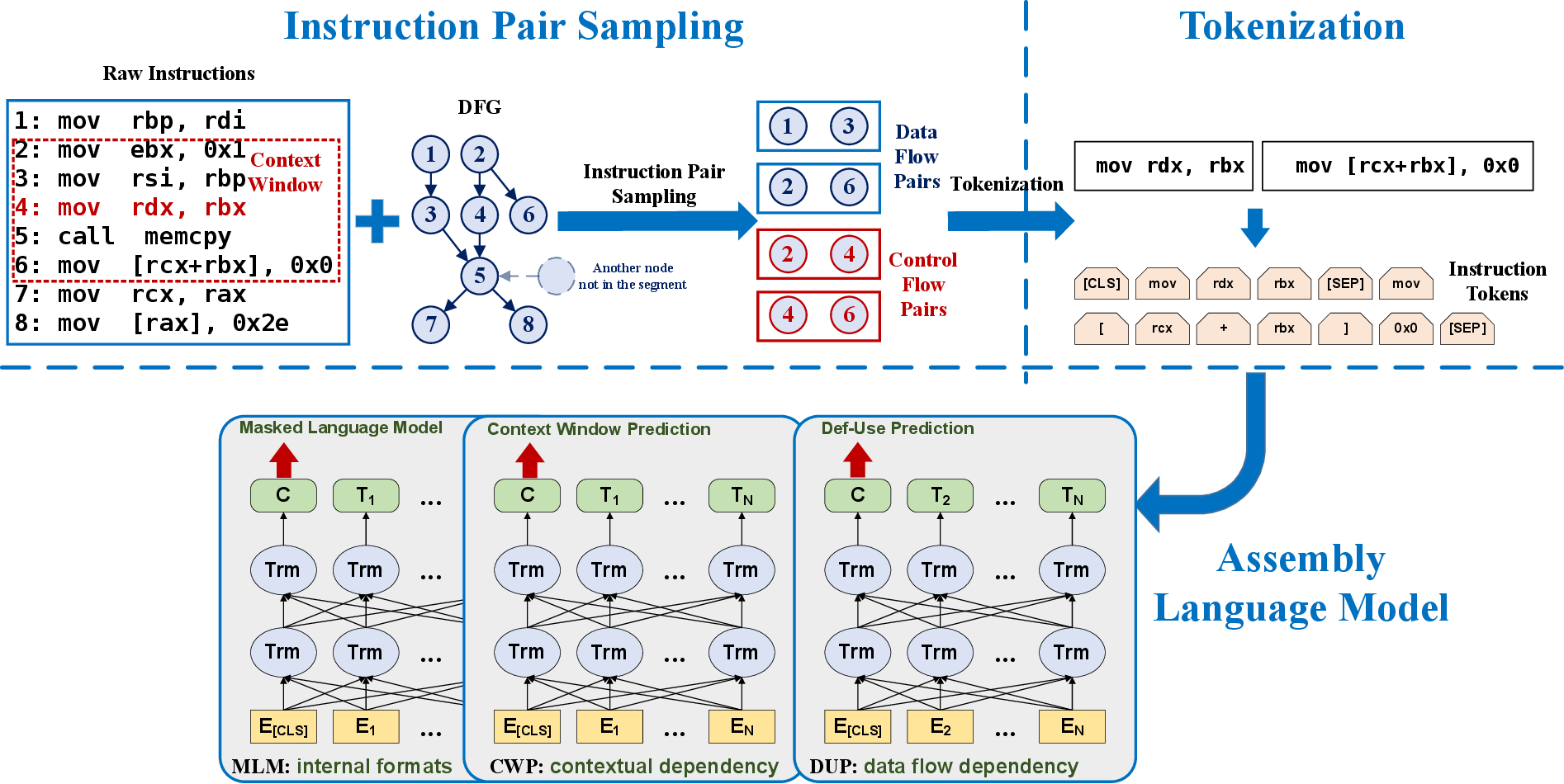}
\caption{System design of \codename. \texttt{Trm} is the transformer encoder unit, \texttt{C} is the hidden state of the first token of the sequence (classification token), ${{T}_{n}}$ ($n=1\ldots N$) are hidden states of other tokens of the sequence}
\label{figure:system}
\end{figure*}

\section{Design of \codename} \label{section: approach}

\subsection{Overview}

To meet the challenges summarized in Section \ref{section:Background}, we propose \codename, a novel instruction embedding scheme that automatically learns a language model for assembly code. \codename is based on BERT~\cite{devlin2019bert}, and incorporates the following important design considerations.

First of all, to capture the complex internal formats of instructions, we use a fine-grained strategy to decompose instructions: we consider each instruction as a sentence and decompose it into basic tokens. 

Then, in order to train the deep neural network to understand the internal structures of instructions, we make use of a recently proposed training task in NLP to train the model: Masked Language Model (MLM)~\cite{devlin2019bert}. This task trains a language model to predict the masked (missing) tokens within instructions.

Moreover, we would like to train this language model to capture the relationships between instructions. To do so, we design a training task, inspired by word2vec~\cite{word2vec2013} and Asm2Vec~\cite{dingasm2vec}, which attempts to infer the word/instruction semantics by predicting two instructions' co-occurrence within a sliding window in control flow. We call this training task Context Window Prediction (CWP), which is based on Next Sentence Prediction (NSP)~\cite{devlin2019bert} in BERT. Essentially, if two instructions $i$ and $j$ fall within a sliding window in control flow and $i$ appears before $j$, we say $i$ and $j$ have a contextual relation. Note that this relation is more relaxed than NSP, where two sentences have to be next to each other. We make this design decision based on our observation described in Section~\ref{subsubsec:context}: instructions may be reordered by compiler optimizations, so adjacent instructions might not be semantically related.

Furthermore, unlike natural language, instruction semantics are clearly documented. For instance, the source and destination operands for each instruction are clearly stated. Therefore, the data dependency (or def-use relation) between instructions is clearly specified and will not be tampered by compiler optimizations. Based on these facts, we design another training task called Def-Use Prediction (DUP) to further improve our assembly language model. Essentially, we train this language model to predict if two instructions have a def-use relation.

\autoref{figure:system} presents the design of \codename. It consists of three components: Instruction Pair Sampling, Tokenization, and Language Model Training. The main component (Assembly Language Model) of the system is based on the BERT model~\cite{devlin2019bert}. After the training process, we use mean pooling of the hidden states of the second last layer of the BERT model as instruction embedding. The Instruction Pair Sampling component is responsible for sampling instruction pairs from binaries based on control flow and def-use relations.

Then, in the second component, the instruction pair is split into tokens. Tokens can be opcode, registers, intermediate numbers, strings, symbols, etc. Special tokens such as strings and memory offsets are encoded and compressed in this step. After that, as introduced earlier, we train the BERT model using the following three tasks: MLM (Masked Language Model), CWP (Context Window Prediction), and Def-Use Prediction (DUP). After the model has been trained, we use the trained language model for instruction embedding generation. In general, the tokenization strategy and MLM will help us address the first challenge in Section \ref{subsection:challenges}, and CWP and DUP can help us address the second challenge.

In Section~\ref{section: Input Generation}, we introduce how we construct two kinds of instruction pairs. In Section~\ref{subsection:Tokenization}, we introduce our tokenization process. Then, we introduce how we design different training tasks to pre-train a comprehensive assembly language model for instruction embedding in Section \ref{section: Architecture}.

\subsection{Input Generation}\label{section: Input Generation}
We generate two kinds of inputs for \codename. First, we disassemble binaries and extract def-use relations. We use Binary Ninja\footnote{https://binary.ninja/} in our implementation, but other disassemblers should work too. With the help of Binary Ninja, we consider dependencies among registers, memory locations, and function call arguments, as well as implicit dependencies introduced by \texttt{\textbf{EFLAGS}}. For each instruction, we retrieve data dependencies of each operand, and identify def-use relations between the instruction and its dependent instructions. Second, we sample instruction pairs from control flow sequences, and also sample instruction pairs based on def-use relations. Instruction pairs from control flow are needed by CWP, while instruction pairs from def-use relations are needed by DUP. MLM can take both kinds of instruction pairs.

\subsection{Tokenization}
\label{subsection:Tokenization}

As introduced earlier, unlike Asm2Vec~\cite{dingasm2vec} which splits an instruction into opcode and up to two operands, we apply a more fine-grained strategy. For instance, given an instruction ``\texttt{\textbf{mov rax, qword [rsp+0x58]}}'', we divide it into ``\texttt{\textbf{mov}}'', ``\texttt{\textbf{rax}}'', ``\texttt{\textbf{qword}}'', ``\texttt{\textbf{[}}'', ``\texttt{\textbf{rsp}}'', ``\texttt{\textbf{+}}'', ``\texttt{\textbf{0x58}}'', and ``\texttt{\textbf{]}}''. In other words, we consider each instruction as a sentence and decompose the operands into more basic elements.

We use the following normalization strategy to alleviate the OOV (Out-Of-Vocabulary) problem caused by strings and constant numbers. For strings, we use a special token \textbf{\texttt{[str]}} to replace them. For constant numbers, if the constants are large (at least five digits in hexadecimal), the exact value is not that useful, so we normalize it with a special token \textbf{\texttt{[addr]}}. If the constants are relatively small (less than four digits in hexadecimal), these constants may carry crucial information about which local variables, function arguments, and data structure fields that are accessed. Therefore we keep them as tokens, and encode them as one-hot vectors.

\ignore{For this part, Reviewer 1 pointed out that: the tokenization seems to rely on proper symbol detection of the binary. This is very much an open problem in binary analysis, and techniques are not perfect to say the least, but the paper glosses over this issue. Some more information here would be good, optimally including an evaluation of how symbolication techniques of different caliber impact the results.}

\subsection{Assembly Language Model} \label{section: Architecture}
In this section we introduce how we apply the BERT model to our assembly language model for instruction embedding, and how we pre-train the model and adopt the model to downstream tasks.

\subsubsection{\codename model}
Our model is based on BERT~\cite{devlin2019bert}, the state-of-the-art PTM in many NLP tasks. The proposed model is a multi-layer bidirectional transformer encoder. Transformer, firstly introduced in 2017~\cite{vaswani2017attention}, is a neural network architecture solely based on multi-head self attention mechanism. In \codename, transformer units are connected bidirectionally and stacked into multiple layers.

\begin{figure}[ht]
\centering
\includegraphics[width=0.95\linewidth]{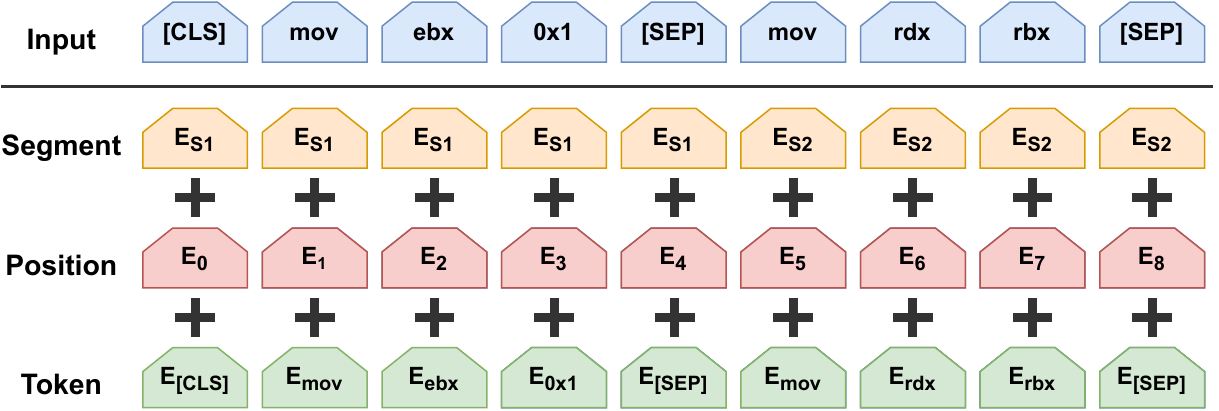}
\caption{Input Representation}
\vspace{-0.1in}
\label{figure:Input Representation}
\end{figure}

We treat each instruction as a sentence and each token as a word. Instructions from control flow and data flow sequences are concatenated and then fed into the BERT model. As shown in \autoref{figure:Input Representation}, the first token of this concatenated input is a special token -- \textbf{\texttt{[CLS]}}, which is used to identify the start of a sequence. Secondly, we use another token \textbf{\texttt{[SEP]}} to separate concatenated instructions. Furthermore, we add position embedding and segment embedding to token embedding, and use this mixed vector as the input of the bi-directional transformer network, as shown in \autoref{figure:Input Representation}. Position embedding represents different positions in the input sequence, while segment embedding distinguishes the first and second instructions. Position embedding and segment embedding will be trained along with token embeddings. These two embeddings can help dynamically adjust token embeddings according to their locations.

\subsubsection{Training task 1: Masked Language Model}
\begin{figure}[ht]
\centering
\includegraphics[width=0.85\linewidth]{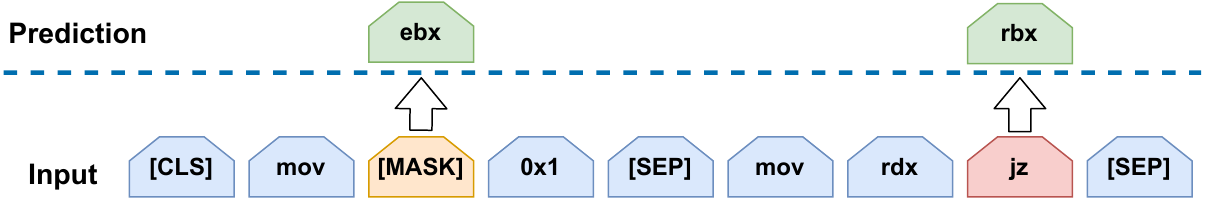}
\caption{Masked Language Model (MLM)}
\vspace{-0.1in}
\label{figure:Masked Language Model}
\end{figure}

The first task we use to pre-train \codename is Masked Language Model (MLM), which was firstly introduced in BERT~\cite{devlin2019bert}. Here is an example shown in \autoref{figure:Masked Language Model}. Assuming that $t_i$ denotes a token and instruction $I=t_1,t_2,t_3,...,t_n$ consists of a sequence of tokens. For a given input instruction $I$, we first randomly select 15\% of the tokens to replace. For the chosen tokens, 80\% are masked by \textbf{\texttt{[MASK]}} (mask-out tokens), 10\% are replaced with another token in the vocabulary (corrupted tokens), and 10\% of the chosen tokens are unchanged. Then, the transformer encoder learns to predict the masked-out and corrupted tokens, and outputs a probability for predicting a particular token $t_i=[MASK]$ with a softmax layer located on the top of the transformer network:
\begin{equation}
p(\hat{t_i}|I)=\frac{exp(w_i\Theta(I)_i)}{\sum_{k=1}^{K}{exp(w_k\Theta(I)_i)}}
\end{equation}
where $\hat{t_i}$ denotes the prediction of $t_i$. $\Theta(I)_i$ is the $i^{th}$ hidden vector of the transformer network $\Theta$ in the last layer, when having $I$ as input. and $w_i$ is weight of label $i$. $K$ is the number of possible labels of token $t_i$. The model is trained with the Cross Entropy loss function:
\begin{equation}
\mathcal{L}_{MLM}=-\sum_{t_i\in m(I)}{\log{p(\hat{t}}|I)} 
\end{equation}
where $m(I)$ denotes the set of tokens that are masked.

\autoref{figure:Masked Language Model} shows an example. Given an instruction pair ``\textbf{\texttt{mov ebx, 0x1; mov rdx, rbx}}'', we first add special tokens \textbf{\texttt{[CLS]}} and \textbf{\texttt{[SEP]}}. Then we randomly select some tokens for replacement. Here we select \textbf{\texttt{ebx}} and \textbf{\texttt{rbx}}. The token \textbf{\texttt{ebx}} is replaced by the \textbf{\texttt{[MASK]}} token (the yellow box). The token \textbf{\texttt{rbx}} is replaced by the token \textbf{\texttt{jz}} (another token in the vocabulary, the red box). Next, we feed this modified instruction pair into the \codename model. The model will make a  prediction for each token. Here we care about the predictions of the yellow and red boxes, which are the green boxes in~\autoref{figure:Masked Language Model}. Only the predictions of those two special tokens are considered in calculating the loss function.

\subsubsection{Training task 2: Context Window Prediction}
We use this training task to capture control flow information. Many downstream tasks \cite{Gemini:ccs17, DeepVSA, zuo2018InnerEYE,Chua17EKLAVYA} rely on the understanding of contextual relations of code sequences in functions or basic blocks. Instead of predicting the whole following sentence (instruction) \cite{DBLP:SutskeverVL14seq2seq,Kiros2015skp-tht}, we perform a binary classification to predict whether the two given instructions co-occur within a context window or not, which makes it a much easier task compared to the whole sentence prediction. However, unlike natural language, control flows do not have strict dependencies and ordering. As a result, strict Next Sentence Prediction (NSP), firstly proposed by BERT~\cite{devlin2019bert}, may not be suitable for capturing contextual information of control flow. To tackle this issue, we extend the context window, i.e., we treat each instruction $w$ steps before and $w$ steps after the target instruction in the same basic block as contextually related. $w$ is the context windows size. In Section~\ref{subsec:context_window_size}, we evaluate the performance of different context window sizes, and pick $w=2$ accordingly.
Given an instruction $I$ and a candidate instruction $I_{cand}$ as input, the candidate instruction can be located in the contextual window of $I$, or a negative sample randomly selected from the dataset. $\hat{y}$ denotes the prediction of this model. The probability that the candidate instruction $I_{cand}$ is a context instruction of $I$ is defined as

\begin{equation}
p(\hat{y}|I, I_{cand}) = \frac{1}{1+{exp(\Theta(I \parallel I_{cand})_{cls})}}
\end{equation}
where $I_{cand} \in \mathbb{C}$, and $\mathbb{C}$ is the candidate set including negative and positive samples. $\Theta_{cls}$ is the first output of the transformer network in the last layer. And ``$\parallel$'' means a concatenation of two instructions. Suppose all instructions belongs to the training set $\mathcal{D}$, then the loss function is:

\begin{equation}
    \mathcal{L}_{CWP}= -\sum_{I \in \mathcal{D}}{\log{p(\hat{y}|I, I_{cand}})}
\end{equation}

\begin{figure}[ht]
\centering
\includegraphics[width=0.85\linewidth]{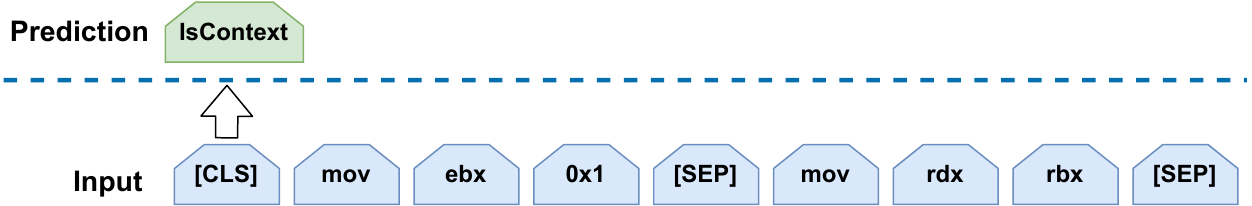}
\caption{Context Window Prediction (CWP)}
\vspace{-0.1in}
\label{figure: Context Window Prediction}
\end{figure}

Here is an example in \autoref{figure: Context Window Prediction}. We use the input mentioned above. We feed the unchanged instruction pairs into the \codename model and pick the first output vector. We use this vector to predict whether the input are located in the same context window or not. In this case, the two instructions are next to each other. Therefore the correct prediction would be ``true''. 

\subsubsection{Training task 3: Def-Use Prediction}
To further improve the quality of our instruction embedding, we need not only control flow information but also data dependency information across instructions.

Sentence Ordering Prediction (SOP), first introduced by Lan et al.~\cite{lan2019albert}, is a very suitable choice. This task can help the \codename model to understand the data relation through DFGs, and we call it Def-Use Prediction (DUP).

Given an instruction pair $I_1$ and $I_2$ as input. And we feed $I_1  \parallel I_2$ as a positive sample and $I_2  \parallel I_1$ as a negative sample. $\hat{y}$ denotes the prediction of this model. The probability that the instruction pair is swapped or not is defined as

\begin{equation}
p(\hat{y}|I_1, I_2) = \frac{1}{1+{exp(\Theta(I_1 \parallel I_2)_{cls})}}
\end{equation}
where $\Theta_{cls}$ is the first output of the transformer network in the last layer. The Cross Entropy loss function is:

\begin{equation}
    \mathcal{L}_{DUP} = -\sum_{I \in \mathcal{D}}{{p(\hat{y}|{{I}_{1}},{{I}_{2}}})}
\end{equation}

\begin{figure}[ht]
\centering
\includegraphics[width=0.85\linewidth]{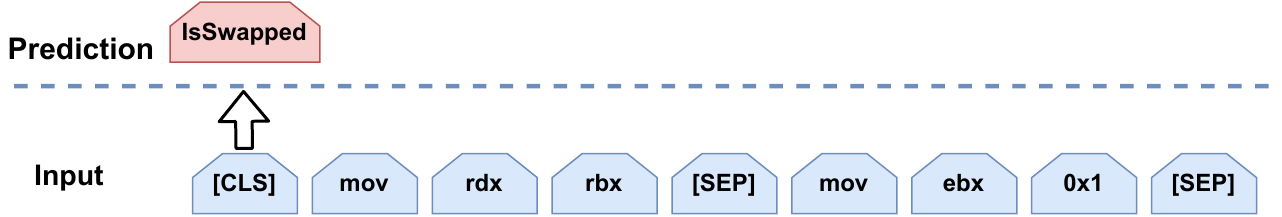}
\caption{Def-Use Prediction (DUP)}
\label{figure: Def-Use Prediction}
\end{figure}

We show an example in~\autoref{figure: Def-Use Prediction}. We still use the instruction pair discussed in~\autoref{figure: Context Window Prediction}, but here we swap the two instructions. So the sequence is ``\textbf{\texttt{[CLS] mov rdx rbx [SEP] mov  ebx 0x1 [SEP]}}''. We feed it into \codename and use the first output vector to predict whether this instruction pair remains unswapped or not. In this case, it should be predicted as ``false'' (which means this pair is swapped).

\medskip\noindent
The loss function of \codename is the combination of three loss functions:
\begin{equation}
    \mathcal{L} = \mathcal{L}_{MLM} + \mathcal{L}_{CWP} + \mathcal{L}_{DUP}
\end{equation}

\subsubsection{Instruction Representation}
\label{subsubsection: Instruction Representation}
The transformer encoder produces a sequence of hidden states as output. There are multiple ways to generate instruction embeddings from the output. For instance, applying a max/mean pooling. We use mean pooling of the hidden states of the second last layer to represent the whole instruction. This design choice has the following considerations. First, the transformer encoder encodes all the input information into the hidden states. A pooling layer is a good way to 
utilize the information encoded by transformer. Second, results in BERT~\cite{devlin2019bert} also suggest that hidden states of previous layers before the last layer have offer more generalizability than the last layer for some downstream tasks. We evaluated different layer configurations and reported the results in Section~\ref{subsec:output_layer}.

\subsubsection{Deployment of the model}
\label{subsubsection: Deployment of the model}

There are two ways of deploying \codename for downstream applications: \textit{instruction embedding generation}, where the pre-trained parameters are frozen, and \textit{fine-tuning}, where the pre-trained parameters can be further adjusted.

In the first way (instruction embedding generation), \codename is used as an off-the-shelf assembly language model to generate high-quality instruction embeddings. Downstream applications can directly use the generated embeddings in their models. Our evaluation results show that \codename without fine-tuning can still outperform existing instruction embedding models such as word2vec and Asm2Vec. This scheme is also very useful when computing resources are limited such as on a lower-end or embedded devices. In this scenario, we can further improve the efficiency by generating a static embedding lookup table in advance. This lookup table contains the embeddings of most common instructions. A trade-off should be made between the model accuracy and the available resources when choosing the lookup table size. A larger lookup table will consume more space but can alleviate the OOV problem (happens when the encountered instruction is not in the table) and improve the accuracy.

In the second way (fine-tuning), \codename is fine-tuned and trained together with the downstream model. This scheme will usually provide extra benefits when enough computing resources and training budget are available. There are several fine-tuning strategies~\cite{qiu2020pre}, e.g., two-stage fine-tuning, multi-task fine-tuning.

\section{Evaluation}

Previous binary analysis studies usually evaluate their approaches by designing specific experiments in an end-to-end manner, since their instruction embeddings are only for individual tasks. In this paper, we focus on evaluating different instruction embedding schemes. To this end, we have designed and implemented an extensive evaluation framework to evaluate \codename and the baseline approaches. Evaluations can be classified into two categories: \textit{intrinsic evaluation} and \textit{extrinsic evaluation}. In the remainder of this section, we first introduce our evaluation framework and experimental configurations, then report and discuss the experimental results.

\subsection{Evaluation Methodology}

\paragraph{Intrinsic Evaluation}
In NLP domain, intrinsic evaluation refers to the evaluations that compare the generated embeddings with human assessments~\cite{Bakarov2018}. Hence, for each intrinsic metric, manually organized datasets are needed. This kind of dataset could be collected either in laboratory on a limited number of examinees or through crowd-sourcing~\cite{liza-grzes-2016-improved} by using web platforms or offline survey~\cite{Bakarov2018}. Unlike the evaluations in NLP domain, programming languages including assembly language (instructions) do not necessarily rely on human assessments. Instead, each opcode and operand in instructions has clear semantic meanings, which can be extracted from instruction reference manuals. Furthermore, debug information generated by different compilers and compiler options can also indicate whether two pieces of code are semantically equivalent. More specifically, we design two intrinsic evaluations: \textit{instruction outlier detection} based on the knowledge of semantic meanings of opcodes and operands from instruction manuals, and \textit{basic block search} by leveraging the debug information associated with source code.

\paragraph{Extrinsic Evaluation}
Extrinsic evaluation aims to evaluate the quality of an embedding scheme along with a downstream machine learning model in an end-to-end manner~\cite{Bakarov2018}. So if a downstream model is more accurate when integrated with instruction embedding scheme A than the one with scheme B, then A is considered better than B. In this paper, we choose three different binary analysis tasks for extrinsic evaluation, i.e., Gemini~\cite{Gemini:ccs17} for \textit{binary code similarity detection}, EKLAVYA~\cite{Chua17EKLAVYA} for \textit{function type signatures inference}, and DeepVSA~\cite{DeepVSA} for \textit{value set analysis}. We obtained the original implementations of these downstream tasks for this evaluation. All of the downstream applications are implemented based on TensorFlow\footnote{https://www.tensorflow.org/}. Therefore we choose the first way of deploying \codename in extrinsic evaluations (see Section~\ref{subsubsection: Deployment of the model}). We encoded all the instructions in the corresponding training and testing datasets and then fed the embeddings into downstream applications.

\subsection{Experimental Setup}

\paragraph{Baseline Schemes and \codename Configurations}
We choose Instruction2Vec, word2vec, and Asm2Vec as baseline schemes. For fair comparison, we set the embedding dimension as 128 for each model. We performed the same normalization method as \codename on word2vec and Asm2Vec. We did not set any limitation on the vocabulary size of Asm2Vec and word2vec. We implemented these baseline embedding models and \codename using PyTorch~\cite{NEURIPS2019_9015}. \codename is based on BERT but has fewer parameters. While in BERT \#$Layers=12$, $Head=12$ and $Hidden\_dimension=768$, we set \#$Layers=12$, $Head=8$, $Hidden\_dimension=128$ in \codename, for the sake of efficiency and training costs. The ratio between the positive and negative pairs in both CWP and DUP is 1:1. 

Furthermore, to evaluate the contributions of three training tasks of \codename, we set up three configurations:
\begin{itemize}
    \item \textbf{\codename-M}:~~ \codename trained with MLM only
    \item \textbf{\codename-MC}: \codename trained with MLM and CWP
    \item \textbf{\codename}:~~~~~~~ \codename trained with MLM, CWP, and DUP
\end{itemize}

\paragraph{Datasets}

To pre-train \codename and evaluate its transferability and generalizability, and evaluate baseline schemes in different downstream applications, we used different binaries from different compilers. The pre-training dataset contains different versions of Binutils\footnote{https://www.gnu.org/software/binutils/}, Coreutils\footnote{https://www.gnu.org/software/coreutils/}, Diffutils\footnote{https://www.gnu.org/software/diffutils/}, and Findutils\footnote{https://www.gnu.org/software/findutils/} on x86-64 platform and compiled with Clang\footnote{https://clang.llvm.org/} and GCC\footnote{https://gcc.gnu.org/} with different optimization levels. The whole pre-training dataset contains \textbf{3,266 binaries} and \textbf{2.25 billion} instructions in total. There are about 2.36 billion positive and negative sample pairs during training. To make sure that training and testing datasets do not have much code in common in extrinsic evaluations, we selected completely different testing dataset from different binary families and compiled by different compilers. Please refer to the following sections for more details about dataset settings.

\paragraph{Hardware Configuration}
All the experiments were conducted on a dedicated server with a Ryzen 3900X CPU@3.80GHz$\times $12, one GTX 2080Ti GPU, 64 GB memory, and 500 GB SSD.

\subsection{Intrinsic Evaluation} \label{section:Intrinsic Evaluation}

\subsubsection{Outlier Detection} \label{section: Outlier word detection}

In this intrinsic evaluation, we randomly create a set of instructions, one of which is an outlier. That is, this instruction is obviously different from the rest of the instructions in this set. To detect this outlier, we calculate the cosine distance between any two instructions' vector representations (i.e., embeddings), and pick whichever is most distant from the rest. We designed two outlier detection experiments, one for opcode outlier detection, and one for operand, to evaluate whether the instruction embeddings are good enough to distinguish different types of opcodes and operands respectively. 


We classify instructions into 12 categories based on their opcode, according to the x86 Assembly Language Reference Manual~\cite{x86manual}. More details about this process can be found in \autoref{table: Types of opcodes} in the Appendix. We prepared 50,000 instruction sets. Each set consists of four instructions from the same opcode category and one instruction from a different category.

\begin{table}[h]
\centering
\small
\caption{Intrinsic Evaluation Results, Avg. denotes the average of accuracy scores, and Stdev. denotes the standard deviation}
\label{table: Instrinsic Evaluations}
\begin{tabular}{lccccc}
\toprule
& \multicolumn{2}{c}{\textbf{\begin{tabular}[c]{@{}c@{}}opcode \\ outlier \end{tabular}}} & \multicolumn{2}{c}{\textbf{\begin{tabular}[c]{@{}c@{}}operand \\ outlier \end{tabular}}} & \multicolumn{1}{c}{\textbf{\begin{tabular}[c]{@{}c@{}}basicblock \\ sim search\end{tabular}}} \\ \cline{2-6} 
\multirow{-3}{*}{\textbf{Model }} & \multicolumn{1}{c}{Avg.} & Stdev.  & \multicolumn{1}{c}{Avg.} & Stdev. & \multicolumn{1}{c}{AUC}\\ \hline
Instruction2Vec &        {0.863} & {0.0529} &        {0.860} & {0.0363}  &        {0.871} \\
word2vec        &        {0.269} & {0.0863} &        {0.256} & {0.0874}  &        {0.842} \\
Asm2Vec         &        {0.865} & {0.0426} &        {0.542} & {0.0238}  &        {0.894} \\ 
\codename-M     &        {0.855} & {0.0333} &        {0.785} & {0.0656}  &        {0.910} \\
\codename-MC    &        {0.870} & {0.0449} &        {0.808} & {0.0435}  &        {0.913} \\
\codename       & \textbf{0.871} & {0.0440} & \textbf{0.944} & {0.0343}  & \textbf{0.922} \\
\bottomrule
\end{tabular}
\end{table}

Similarly, we classify instructions based on their operands. \autoref{table: Types of operands} in the Appendix provides details about this process. Essentially, we classify operand lists, according to the number of operands as well as the operand types. We created another 50,000 sets of instructions covering 10 categories, and each set contains four instructions coming from the same category, and one from a different category.

\begin{figure}[ht]
\centering
\includegraphics[width=0.7\linewidth]{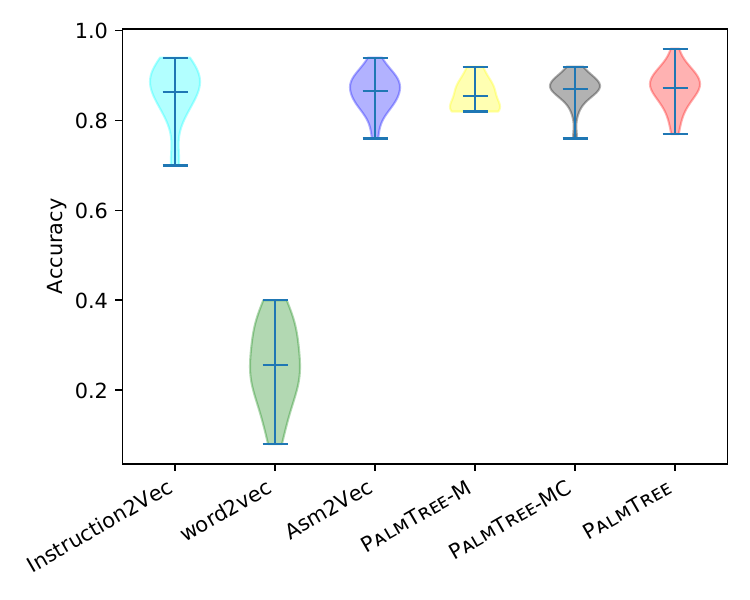}
\vspace{-0.2in}
\caption{Accuracy of Opcode Outlier Detection}
\label{figure:outlier_detection: opcode}
\end{figure}

\begin{figure}[ht]
\centering
\includegraphics[width=0.7\linewidth]{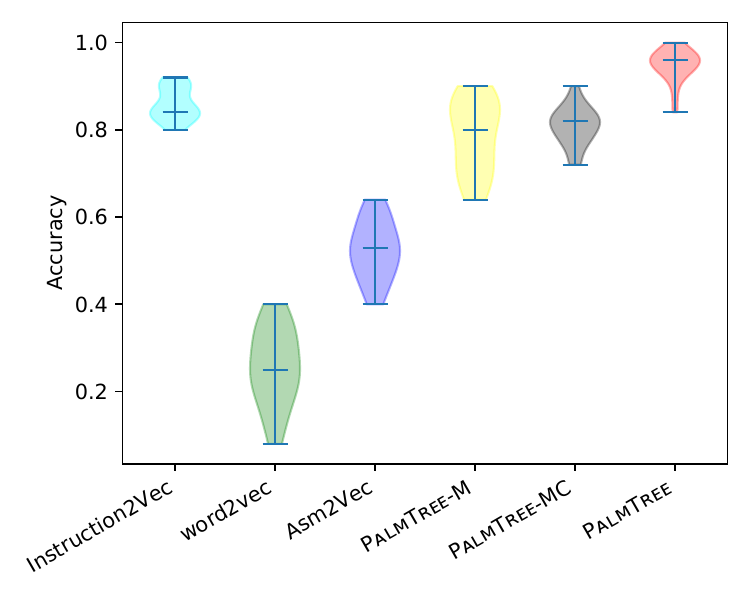}
\vspace{-0.2in}
\caption{Accuracy of Operands Outlier Detection}
\label{figure:outlier_detection: operand}
\end{figure}

The first and second columns of \autoref{table: Instrinsic Evaluations} present the accuracy distributions for opcode outlier detection and operand outlier detection respectively. We can make the following observations: (1) word2vec performs poorly in both experiments, because it does not take into account the instruction internal structures; (2) Instruction2Vec, as a manually-designed embedding, performs generally well in both experiments, because this manual design indeed takes different opcodes and operands into consideration; (3) Asm2Vec performs slightly better than Instruction2Vec in opcode outlier detection, but considerably worse in operand outlier detection, because its modeling for operands is not fine-grained enough; (4) Even though \codename-M and \codename-MC do not show obvious advantages over Asm2Vec and Instruction2Vec, \codename has the best accuracy in both experiments, which demonstrate that this automatically learned representation can sufficiently capture semantic differences in both opcodes and operands; and (5) All the three pre-training tasks contribute positively to \codename in both outlier detection experiments. Particularly, the DUP training task considerably boots the accuracy in both experiments, demonstrating that the def-use relations between instructions indeed help learn the assembly language model. A complete result of outlier detection can be found in \autoref{figure:outlier_detection: opcode} and \autoref{figure:outlier_detection: operand}.

\subsubsection{Basic Block Search} \label{section: Basic block level similarity search}

In this intrinsic evaluation, we compute an embedding for each basic block (a sequence of instructions with only one entry and one exit), by averaging the instruction embeddings in it. Given one basic block, we use its embedding to find semantically equivalent basic blocks based on the cosine distance between two basic block embeddings.

We use {\tt openssl-1.1.0h} and {\tt glibc-2.29.1} as the testing set, which is not included in our training set. We compile them with O1, O2, and O3 optimization levels. We use the same method used in DeepBinDiff~\cite{duan2020deepbindiff}, which relies on the debug information from the program source code as the ground truth.

\begin{figure}[t]
\centering
\includegraphics[width=0.7\linewidth]{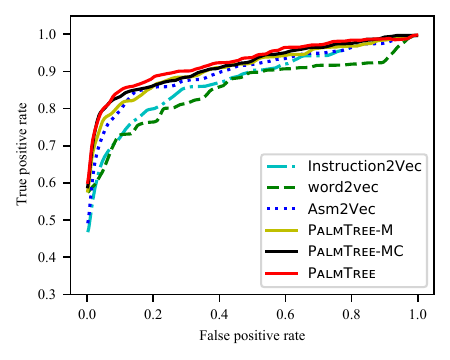}
\vspace{-0.2in}
\caption{ROC curves for Basic Block Search}
\label{figure:bb_levl_sim_search}
\end{figure}

\autoref{figure:bb_levl_sim_search} shows the ROC curves of Instruction2Vec, word2vec, Asm2Vec, and \codename for basic block search. \autoref{table: Instrinsic Evaluations} further lists the AUC (Area Under the Curve) score for each embedding scheme. We can observe that (1) word2vec, once again, has the worst performance; (2) the manually-designed embedding scheme, Instruction2Vec, is even better than word2vec, an automatically learned embedding scheme; (3) Asm2Vec performs reasonably well, but still worse than three configurations of \codename; and (4) The three \codename configurations have better AUC than other baselines, while consecutive performance improvements are observed.

\begin{center}
\fcolorbox{black}{gray!15}{\parbox{.95\linewidth}{\codename ranks the first in all intrinsic evaluation experiments, demonstrating the strength of the automatically learned assembly language model. And the performance improvements between different \codename configurations show positive contributions of individual training tasks.

}}
\end{center}

\subsection{Extrinsic Evaluation} \label{section: Extrinsic Evaluation}
An extrinsic evaluation reflects the ability of an instruction embedding model to be used as an input of downstream machine learning algorithms for one or several specific tasks \cite{Bakarov2018}. As introduced earlier, we select three downstream tasks in binary analysis field, which are binary code similarity detection, function type signature analysis, and value set analysis.

\subsubsection{Binary Code Similarity Detection}\label{subsubsec:gemini}

\begin{figure}[th]
\centering
\includegraphics[width=0.9\linewidth]{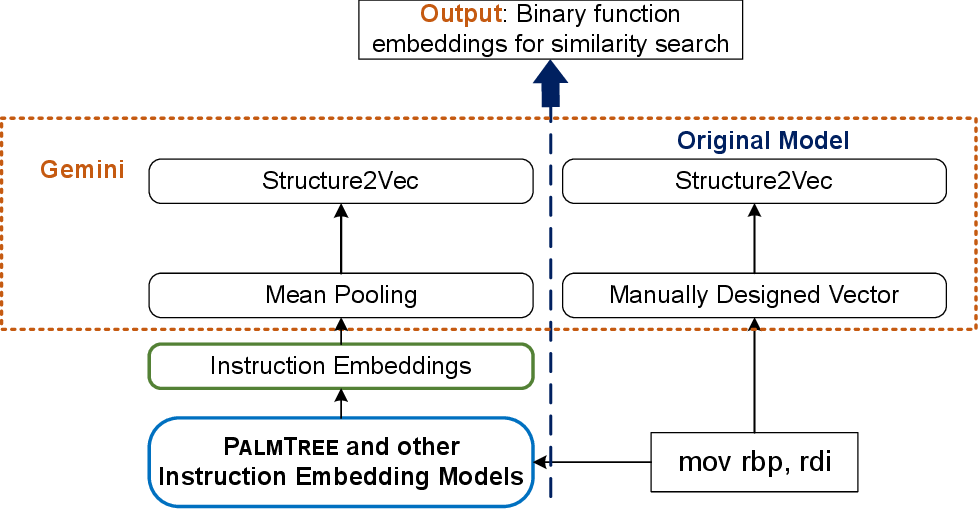}
\caption{Instruction embedding models and the downstream model Gemini}
\label{figure:BITE and Gemini}
\end{figure}

Gemini~\cite{Gemini:ccs17} is a neural network-based approach for cross-platform binary code similarity detection. The model is based on Structure2Vec~\cite{dai2016s2v} and takes ACFG (Attributed Control Flow Graph) as input. In an ACFG, each node is a manually formed feature vector for each basic block. \autoref{table: Attributes of Basic Blocks} shows the attributes (i.e., features) of a basic block in the original implementation.

\begin{table}[h]
\centering
\footnotesize
\caption{Attributes of Basic Blocks in Gemini~\cite{Gemini:ccs17}}
\label{table: Attributes of Basic Blocks}
\begin{tabular}{ll}
\toprule
\textbf{Type} & \textbf{Attribute name} \\ \midrule
\multirow{6}{*}{\textbf{Block-level attributes}} & String Constants \\
 & Numeric Constants \\
 & No. of Transfer Instructions \\
 & No. of Calls \\
 & No. of Instructions \\
 & No. of Arithmetic Instructions \\ 
\midrule
\multirow{2}{*}{\textbf{Inter-block attributes}} & No. of offspring \\
 & Betweenness \\ 
\bottomrule
\end{tabular}
\end{table}

In this experiment, we evaluate the performance of Gemini, when having Instruction2Vec, word2vec, Asm2Vec, \codename-M, \codename-MC, and \codename as input, respectively. Moreover, we also used one-hot vectors with an embedding layer as a kind of instruction embedding (denoted as ``one-hot'') as another baseline. The embedding layer will be trained along with Gemini. \autoref{figure:BITE and Gemini} shows how we adopt different instruction embedding models to Gemini.  Since Gemini takes a feature vector for each basic block, we use mean pooling to generate basic block embeddings based on embeddings of the instructions in the corresponding basic block. The architectures of our modified model and the original model are both shown in \autoref{figure:BITE and Gemini}. We also included its original basic block features as an additional baseline (denoted as ``Gemini'') for comparison.

\begin{figure}[ht]
\centering
\includegraphics[width=0.75\linewidth]{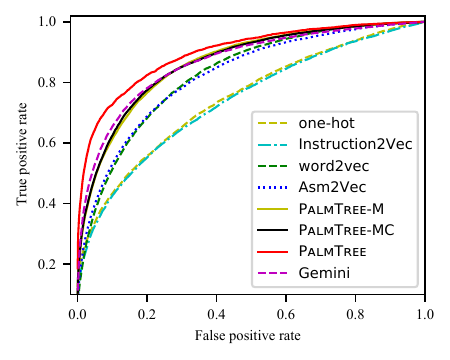}
\vspace{-0.2in}
\caption{ROC curves of Gemini}
\label{figure: ROC cuvre of Gimini}
\end{figure}

The accuracy of the original Gemini is reported to be very high (with an AUC of 0.971). However, this might be due to overfitting, since the training and testing sets are from OpenSSL compiled by the same compiler Clang. To really evaluate the generalizability (i.e., the ability to adapt to previously unseen data) of the trained models under different inputs, we use {\tt binutils-2.26}, {\tt binutils-2.30}, and {\tt coreutils-8.30} compiled by Clang as training set (237 binaries in total), and used {\tt openssl-1.1.0h}, {\tt openssl-1.0.1}, and {\tt glibc-2.29.1} compiled by GCC as testing set (14 binaries). In other words, the training and testing sets are completely different and the compilers are different too.

\begin{table}[h]
\centering
\small
\caption{AUC values of Gemini}
\label{table:AUC in Gemini}
\begin{tabular}{lc|lc}
\toprule

\textbf{Model}              & \textbf{AUC} & \textbf{Model}  & \textbf{AUC}  \\ \midrule
one-hot            & 0.745 & Gemini & 0.866  \\
Instruction2Vec    & 0.738 & \codename-M          & 0.864 \\
word2vec           & 0.826 & \codename-MC     & 0.866 \\
Asm2Vec            & 0.823 & \codename & \textbf{0.921} \\
\bottomrule
\end{tabular}
\end{table}

\autoref{table:AUC in Gemini} gives the AUC values of Gemini when different models are used to generate its input. Figure \ref{figure: ROC cuvre of Gimini} shows the ROC curves of Gemini when different instruction embedding models are used. Based on \autoref{table:AUC in Gemini}, we can make the following observations:

\begin{enumerate}[label=(\arabic*), ref=\arabic*]
    \item  Although the original paper~\cite{Gemini:ccs17} reported very encouraging performance of Gemini, we can observe that the original Gemini model does not generalize very well to completely new testing data. 
    
    \item The manually designed embedding schemes, Instruction2Vec and one-hot vector, perform poorly, signifying that manually selected features might be only suitable for specific tasks.

    \item Despite that the testing set is considerably different from the training set, \codename can still perform reasonably well and beat the remaining schemes, demonstrating that \codename can substantially boost the generalizability of downstream tasks.

    \item All the three pre-training tasks contribute to the final model (\codename) for Gemini. However, both \codename-M and \codename-MC do not show obvious advantages over other baselines, signifying that only the complete \codename with the three training tasks can generate better embeddings than previous approaches in this downstream task.

\end{enumerate}

\subsubsection{Function Type Signature Inference}

\begin{figure}[ht]
\centering
\includegraphics[width=0.95\linewidth]{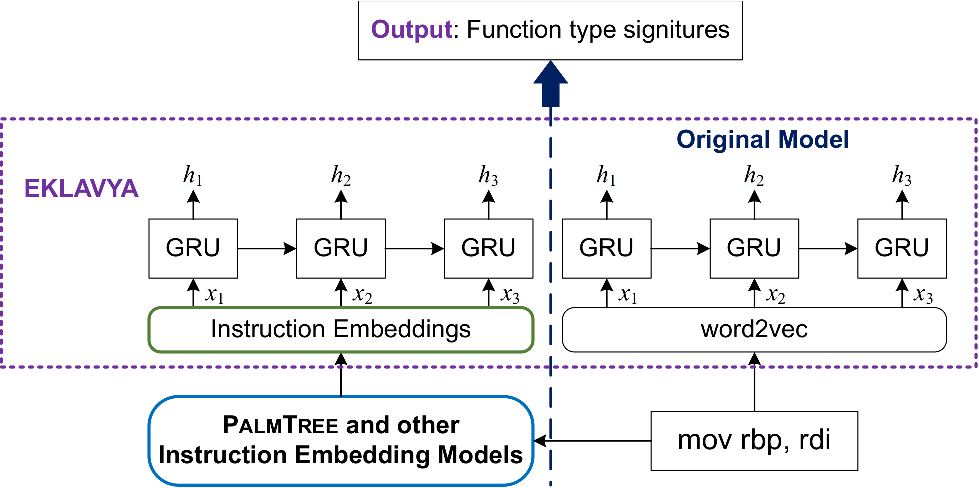}
\caption{Instruction embedding models and EKLAVYA}
\label{figure:BITE and EKLAVYA}
\end{figure}

Function type signature inference is a task of inferring the number and primitive types of the arguments of a function. To evaluate the quality of instruction embeddings in this task, we select EKLAVYA, an approach proposed by Chua et al.~\cite{Chua17EKLAVYA}. It is based on a multi-layer GRU (Gated Recurrent Unit) network and uses word2vec as the instruction embedding method. According to the original paper, word2vec was pre-trained with the whole training dataset. Then, they trained a GRU network to infer function type signatures.

In this evaluation, we test the performances of different types of embeddings using EKLAVYA as the downstream application. Since the original model is not an end-to-end model, we do not need an embedding layer between instruction embeddings and the GRU network. We replaced the original word2vec in EKLAVYA with one-hot encoding, Instruction2Vec, Asm2Vec, \codename-M, \codename-MC, and \codename, as shown in \autoref{figure:BITE and EKLAVYA}. 

Similarly, in order to evaluate the generalizability of the trained downstream models, we used very different training and testing sets (the same datasets described in Section~\ref{subsubsec:gemini}).

\begin{figure}[ht]
\centering
\includegraphics[width=0.8\linewidth]{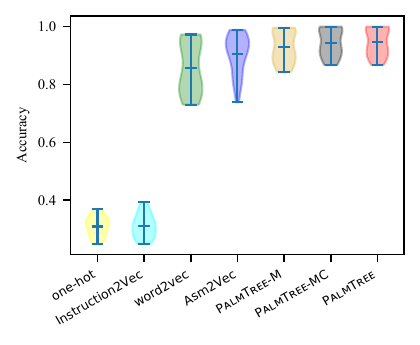}
\vspace{-0.2in}
\caption{Accuracy of EKLAVYA}
\label{figure: accuracy of Eklavya on testing dataset}
\end{figure}

\begin{table}[h]
\centering
\small
\caption{Accuracy and Standard Deviation of EKLAVYA}
\label{table:Accuracy of EKLAVYA}
\begin{tabular}{lcc}
\toprule

\textbf{Model}              & \textbf{Accuracy} & \textbf{Standard Deviation} \\ \midrule
one-hot            & 0.309 & 0.0338 \\
Instruction2Vec    & 0.311 & 0.0407 \\ 
word2vec           & 0.856 & 0.0884 \\
Asm2Vec            & 0.904 & 0.0686 \\
\codename-M          & 0.929 & 0.0554 \\
\codename-MC     & 0.943 & 0.0476 \\
\codename & \textbf{0.946} & 0.0475\\ \bottomrule
\end{tabular}
\end{table}

\autoref{table:Accuracy of EKLAVYA} and \autoref{figure: accuracy of Eklavya on testing dataset} presents the accuracy of EKLAVYA on the testing dataset. \autoref{figure: loss value of Eklavya}, and \autoref{figure: accuracy of Eklavya in training} in the Appendix shows the loss value and accuracy of EKLAVYA during training and testing. From the results we can make the following observations:

\begin{enumerate}[label=(\arabic*), ref=\arabic*]

\item \codename and Asm2Vec can achieve higher accuracy than word2vec, which is the original choice of EKLAVYA. 

\item \codename has the best accuracy on the testing dataset, demonstrating that EKLAVYA when fed with \codename as instruction embeddings can achieve the best generalizability. Moreover, CWP contributes more (see \codename-MC), which implies that control-flow information plays a more significant role in EKLAVYA.

\item Instruction2Vec performs very poorly in this evaluation, signifying that, when not done correctly, manual feature selection may disturb and mislead a downstream model.

\item The poor results of one-hot encoding show that a good instruction embedding model is indeed necessary. At least in this task, it is very difficult for the deep neural network to learn instruction semantic through end-to-end training.

\end{enumerate}

\begin{figure}[ht]
\centering
\includegraphics[width=0.95\linewidth]{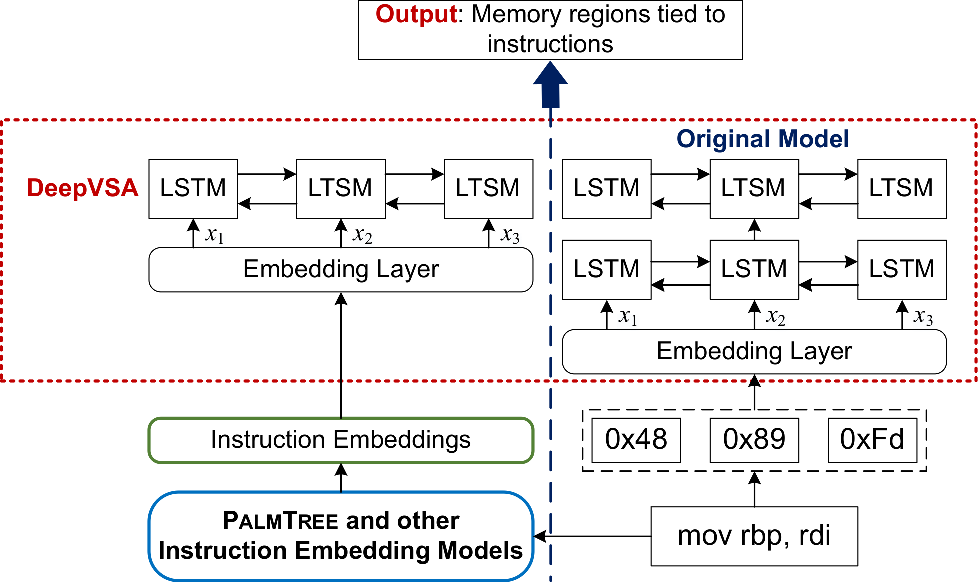}
\caption{Instruction embedding models and the downstream model DeepVSA}
\label{figure:BITE and DeepVSA}
\end{figure}

\subsubsection{Value Set Analysis}

DeepVSA~\cite{DeepVSA} makes use of a hierarchical LSTM network to conduct a coarse-grained value set analysis, which characterizes memory references into regions like global, heap, stack, and other. 
It feeds instruction raw bytes as input into a multi-layer LSTM network to generate instruction embeddings. It then feeds the generated instruction representations into another multi-layer bi-directional LSTM network, which is supposed to capture the dependency between instructions and eventually predict the memory access regions.

In our experiment, we use different kinds of instruction embeddings to replace the original instruction embedding generation model in DeepVSA. We use the original training and testing datasets of DeepVSA and compare prediction accuracy of different kinds of embeddings. The original datasets contain raw bytes only, thus we need to disassemble these raw bytes. After that we tokenize and encode these disassembled instructions for training and testing. We add an embedding layer before the LSTM network to further adjust instruction embeddings, as shown in \autoref{figure:BITE and DeepVSA}.

We use part of the dataset provided by the authors of DeepVSA. The whole dataset provided by the authors has 13.8 million instructions for training and 10.1 million for testing.  Our dataset has 9.6 million instructions for training and 4.8 million for testing, due to the disassembly time costs. As explained in their paper~\cite{DeepVSA}, their dataset also used Clang and GCC as compilers and had no overlapping instructions between the training and testing datasets. 

\begin{figure}[ht]
\centering
\includegraphics[width=0.75\linewidth]{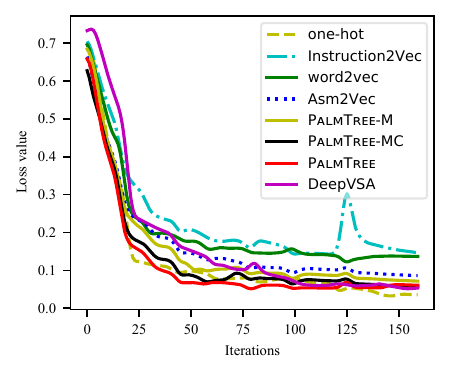}
\vspace{-0.2in}
\caption{Loss value of DeepVSA during training}
\label{figure: loss value of DeepVsa}
\end{figure}

\begin{table*}[ht]
\centering
\small
\caption{Results of DeepVSA}
\begin{tabular}{l|ccc|ccc|ccc|ccc}
\toprule
\multirow{2}{*}{\textbf{Embeddings}} & \multicolumn{3}{c|}{\textbf{Global}} & \multicolumn{3}{c|}{\textbf{Heap}} & \multicolumn{3}{c|}{\textbf{Stack}} & \multicolumn{3}{c}{\textbf{Other}} \\ \cline{2-13} 
 & P & R & F1 & P & R & F1 & P & R & F1 & P & R & F1 \\ \hline
one-hot  & 0.453 & 0.670 & 0.540 & 0.507 & {\bf 0.716} & 0.594 & 0.959 & 0.866 & 0.910 & 0.953 & 0.965 & 0.959 \\
Instruction2Vec & 0.595 & 0.726 & 0.654 & 0.512 & 0.633 & 0.566 & 0.932 & 0.898 & 0.914 & 0.948 & 0.946 & 0.947 \\
word2vec & 0.147 & 0.535 & 0.230 & 0.435 & 0.595 & 0.503 & 0.802 & 0.420 & 0.776 & 0.889 & 0.863 & 0.876 \\
Asm2Vec & 0.482 & 0.557 & 0.517 & 0.410 & 0.320 & 0.359 & 0.928 & 0.894 & 0.911 & 0.933 & 0.964 & 0.948 \\
DeepVSA & {\bf 0.961} & 0.738 & 0.835 & 0.589 & 0.580 & 0.584 &  0.974 & 0.917 & 0.944 & 0.943 & 0.976 & 0.959 \\ 
\codename-M & 0.845 & 0.732 & 0.784 & 0.572 & 0.625 & 0.597 & 0.963 & 0.909 & 0.935 & 0.956 & 0.969 & 0.962 \\
\codename-MC & 0.910 & 0.755 & 0.825 & {\bf0.758} & 0.675 & 0.714 & 0.965 & 0.897 & 0.929 & 0.958 & {\bf 0.988} & {\bf 0.972} \\
\codename & 0.912 & {\bf 0.805} & {\bf 0.855} &  0.755 & 0.678 & {\bf 0.714} & {\bf 0.974} & {\bf 0.929} & {\bf0.950} & {\bf 0.959} & 0.983 & 0.971 \\

\bottomrule
\end{tabular}
\label{table: result of DeepVSA}
\end{table*}

\autoref{table: result of DeepVSA} lists the experimental results. We use Precision (P), Recall (R), and F1 scores to measure the performance. \autoref{figure: loss value of DeepVsa} depicts the loss values of DeepVSA during training, when different instruction embedding schemes are used as its input.
From these results, we have the following observations:

\begin{enumerate}[label=(\arabic*), ref=\arabic*]

\item \codename has visibly better results than the original DeepVSA and the other baselines in Global and Heap, and has slightly better results in Stack and Other since other baselines also have scores greater than 0.9. 

\item The three training tasks of \codename indeed contribute to the final result. It indicates that \codename indeed captures the data flows between instructions. In comparison, the other instruction embedding models are unable to capture data dependency information very well. 

\item \codename converged faster than original DeepVSA (see \autoref{figure: loss value of DeepVsa}), indicating that instruction embedding model can accelerate the training phase of downstream tasks.

\end{enumerate}

\begin{center}
\fcolorbox{black}{gray!15}{\parbox{.95\linewidth}{
\codename outperforms the other instruction embedding approaches in each extrinsic evaluation. Also, \codename can speed up training and further improve downstream models by providing high-quality instruction embeddings. In contrast, word2vec and Instruction2Vec perform poorly in all the three downstream tasks, showing that the poor quality of an instruction embedding will adversely affect the overall performance of downstream applications.}}
\end{center}

\subsection{Runtime Efficiency}

In this section, we conduct an experiment to evaluate runtime efficiencies of \codename and baseline approaches. First, we test the runtime efficiencies of different instruction embedding approaches. 
Second, we test the runtime efficiency of \codename when having different embedding sizes. We use 64, 128, 256, and 512 as embedding sizes, while 128 is the default setting. In the transformer encoder of \codename, the width of each feed-forward hidden layer is fixed and related to the size of the final output layer, which is 4 times of the embedding size ~\cite{lan2019albert}. We use \texttt{Coreutils-8.30} as the dataset. It includes 107 binaries and 1,006,169 instructions. We disassembled the binaries with Binary Ninja and feed them into the baseline models. Due to the limitation of GPU memory, we treated 5,000 instructions as a batch.

\begin{table}[h]
\centering
\small
\caption{Efficiency of \codename and baselines}
\label{table: Efficiency}
\begin{tabular}{lrr}
\toprule
\textbf{embedding size} & \multicolumn{1}{l}{\textbf{encoding time}} & \multicolumn{1}{l}{\textbf{throughput (\#ins/sec)}} \\ \midrule
Instruction2vec             & 6.684              & 150,538                            \\
word2vec                    & 0.421            &  2,386,881                         \\
Asm2Vec                     & 17.250             & 58,328                            \\
\codename-64                    & 41.682              & 24,138                           \\
\textbf{\codename-128}                   & 70.202             & 14,332                            \\
\codename-256                   & 135.233             & 7,440  \\
\codename-512   & 253.355 & 3,971\\ 
\bottomrule
\end{tabular}
\end{table}

\autoref{table: Efficiency} shows the encoding time and throughput of different models when encoding the 107 binaries in \texttt{Coreutils-8.30}. From the results, we can make several observations. First, \codename is much slower than previous embedding approaches such as word2vec and Asm2Vec. This is expected, since \codename has a deep transformer network. However, with the acceleration of the GPU, \codename can finish encoding the 107 binaries in about 70 seconds, which is acceptable. Furthermore, as an instruction level embedding approach, \codename can have an embedding lookup table as well to store some frequently used embeddings. This lookup table works as fast as word2vec and can further boost the efficiency of \codename. Last but not least, from the results we observed that it would be 1.7 to 1.9 times slower when doubling the embedding size. 

\subsection{Hyperparameter Selection}

To further study the influences of different hyperparameter configurations of \codename, we trained \codename with different embedding sizes (64, 128, 256, and 512) and different context window sizes (1, 2, 3, and 4). We also evaluated different output layer configurations when generating instruction embeddings. Interested readers are referred to the Appendix for more details.

\section{Related Work}

\paragraph{Representation Learning in NLP}
Over the past several years, representation learning techniques have made significant impacts in NLP domain. Neural Network Language Model (NNLM)~\cite{NNLM2003} is the first work that used neural networks to model natural language and learn distributed representations for words. In 2013, Mikolov et al. introduced word2vec and proposed Skip-gram and Continuous Bag-Of-Words (CBOW) models~\cite{word2vec2013}. The limitation of word2vec is that its embedding is frozen once trained, while words might have different meanings in different contexts. To address this issue, Peters et al. introduced ELMo~\cite{Peters2018elmo}, which is a deep bidirectional language model. In this model, word embeddings are generated from the entire input sentence, which means that the embeddings can be dynamically adjusted according to different contextual information.


In 2017, Vaswani et al. introduced transformer~\cite{vaswani2017attention} to replace the RNN networks (e.g., LSTM). Devlin et al. proposed BERT~\cite{devlin2019bert} in 2019, which is a bi-directional transformer encoder. They designed the transformer network using a full connected architecture, so that the model can leverage both forward and backward information. Clark et al.~\cite{clark2020electra} proposed ELECTRA and further improved BERT by using a more sample-efficient pre-training task called \textit{Replaced Token Detection}. This task is an adversarial learning process~\cite{goodfellow2014generative}.

\paragraph{Representation Learning for Instructions}
Programming languages, including low level assembly instructions, have clear grammar and syntax, thus can be treated as natural language and be processed by NLP models. 


Instruction representation plays a significant role in binary analysis tasks. Many techniques have been proposed in previous studies. Instruction2Vec~\cite{Lee2017ins2vec} is a manually designed instruction representation approach. InnerEye~\cite{zuo2018InnerEYE} uses Skip-gram, which is one of the two models of word2vec~\cite{word2vec2013}, to encode instructions for code similarity search. Each instruction is treated as a word while a code snippet as a document.  Massarelli et al.~\cite{Massarelli2019Safe} introduced an approach for function-level representation learning, which also leveraged word2vec to generate instruction embeddings. DeepBindiff~\cite{duan2020deepbindiff} also used word2vec to generate representations for instructions with the purpose of matching basic blocks in different binaries. Unlike InnerEye, they used word2vec to learn token embeddings and generate instruction embeddings by concatenating vectors of opcode and operands.


Although word2vec has been widely used in instruction representation learning. It has the following shortcommings: first, using word2vec at the instruction level embedding will lose internal information of instructions; on the other hand, using word2vec at the token level may fail to capture instruction level semantics. Second, the model has to handle the OOV problem. InnerEye~\cite{zuo2018InnerEYE} and DeepBindiff~\cite{duan2020deepbindiff} provided good practices by applying normalization. However, normalization also results in losing some important information. Asm2Vec~\cite{dingasm2vec} generates embeddings for instructions and functions simultaneously by using the PV-DM model~\cite{le2014distributed}. Unlike previous word2vec based approaches, Asm2Vec exploits a token level language model for training and did not have the problem of breaking the boundaries of instructions, which is a problem of token level word2vec models. Coda~\cite{fu2019coda} is a neural program decompiler based on a Tree-LSTM autoencoder network. It is an end-to-end deep learning model which was specifically designed for decompilation. It cannot generate generic representations for instructions, thus cannot meet our goals.

\paragraph{Representation Learning for Programming Languages}
NLP techniques are also widely used to learn representations for programming languages. Harer et al.~\cite{harer2018automated} used word2vec to generate token embeddings of C/C++ programs for vulnerability prediction. The generated embeddings are fed into a TextCNN network for classification. Li et al.~\cite{li2019improving} introduced a bug detection technique using word2vec to learn token (node) embedding from Abstract Syntax Tree (AST). Ben-Nun et al.~\cite{NIPS2018_7617} introduced a new representation learning approach for LLVM IR in 2018. They generated conteXtual Flow Graph (XFG) for this IR, which leverages both data dependency and control flow. Karampatsis et al.~\cite{karampatsis2020big} proposed a new method to reduce vocabulary size of huge source code dataset. They introduced word splitting, subword splitting with Byte Pair Encoding (BPE)~\cite{Sennrich2015bpe} cache, and dynamic adaptation to solve the OOV problem in source code embedding.

\section{Discussion}
In this paper, we focus on training an assembly language model for one instruction set or one architecture. We particularly evaluated x86. The technique described here can be applied to other instruction sets as well, such as ARM and MIPS.

However, in this paper, we do not intend to learn a language model across multiple CPU architectures. Cross-architecture means that semantically similar instructions from different architectures can be mapped to near regions in the embedded space. Cross-architecture assembly language model can be very useful for cross-architecture vulnerability/bug search. We leave it as a future work.

It is worth noting that instead of feeding a pair of instructions into \codename, we can also feed code segment pairs or even basic block and function pairs, which may better capture long-term relations between instructions (currently we use sampling in the context window and data flow graph to capture long-term relations) and has a potential to further improve the performance of \codename. We leave this as a future work.

\section{Conclusion}

In this paper, we have summarized the unsolved problems and existing challenges in instruction representation learning. To solve the existing problems and capture the underlying characteristics of instruction, we have proposed a pre-trained assembly language model called \codename for generating general-purpose instruction embeddings.

\codename can be pre-trained by performing self-supervised training on large-scale unlabeled binary corpora. \codename is based on the BERT model but pre-trained with newly designed training tasks exploiting the inherent characteristics of assembly language. More specifically, we have used the following three pre-training tasks to train \codename: MLM (Masked Language Model), CWP (Context Window Prediction), and DUP (Def-Use Prediction). We have designed a set of intrinsic and extrinsic evaluations to systematically evaluate \codename and other instruction embedding models. Experimental results show that \codename has the best performance in intrinsic evaluations compared with the existing models. In extrinsic evaluations that involve several downstream applications, \codename outperforms all the baseline models and also significantly improves downstream applications' performance. We conclude that \codename can effectively generate high-quality instruction embedding which is helpful for different downstream binary analysis tasks.

\section{Acknowledgement}
We would like to thank the anonymous reviewers for their helpful and constructive comments. This work was supported in part by National Science Foundation under grant No. 1719175, and Office of Naval Research under Award No. N00014-17-1-2893. Any opinions, findings, and conclusions or recommendations expressed in this paper are those of the authors and do not necessarily reflect the views of the funding agencies.

\bibliographystyle{ACM-Reference-Format}
\bibliography{reference}

\appendix

\section{Opcode and Operand Types for Outlier Detection}

\autoref{table: Types of opcodes} shows how we categorize different opcodes by referring to \cite{x86manual}. \autoref{table: Types of operands} shows how we categorize different operand types. The first column shows the type of operands combination. ``none'' means the instruction has no operand, such as \texttt{\textbf{retn}}. ``tri'' means the instruction has three operands. The other ones are instructions that have two operands. For instance, ``reg-reg'' means both operands are registers. The type of each operand has been listed in the second and third columns.

\begin{table}[ht]
\small
\centering
\caption{Types of Opcodes}
\label{table: Types of opcodes}
\begin{tabular}{p{2.5cm}<{\centering}p{3.5cm}}
\toprule
\textbf{Types} & \textbf{Opcodes} \\ \hline
Data Movement &  mov, push, pop, cwtl, cltq, cqto, cqtd\\\hline
Unary Operations & inc, dec, neg, not \\ \hline
Binary Operations & lea, leaq, add, sub,imul, xor, or, and \\\hline
Shift Operations & sal, sar, shr, shl \\\hline
Special Arithmetic Operations & imulq, mulq, idivq, divq \\\hline
Comparison and Test Instructions & cmp, test \\ \hline
Conditional Set Instructions &  sete, setz, setne, setnz, sets, setns, setg, setnle,setge, setnl, setl, setnge,setle, setng, seta, setnbe,  setae, setnb, setbe, setna \\\hline
Jump Instructions & jmp, je, jz, jne, jnz, js, jns, jg, jnle, jge, jnl, jl jnge, jle, jng, ja, jnbe, jae, jnb, jb, jnae, jbe, jna \\ \hline
Conditional Move Instructions & cmove, cmovz, cmovne, cmovenz, cmovs, cmovns, cmovg, cmovnle, cmovge, cmovnl, cmovnge, cmovle, cmovng, cmova, cmovnbe, cmovae, cmovnb, cmovb, cmovnae, cmovbe, cmovna \\ \hline
Procedure Call Instructions & call, leave, ret, retn \\\hline
String Instructions & cmps, cmpsb, cmpsl, cmpsw, lods, lodsb, lodsl, lodsw,mov, movsb, movsl, movsw\\ \hline
Floating Point Arithmetic & fabs, fadd, faddp, fchs, fdiv, fdivp, fdivr, fdivrp, fiadd, fidivr, fimul, fisub, fisubr, fmul, fmulp, fprem, fpreml,frndint, fscale, fsqrt, fsub,fsubp, fsubr, fsubrp, fxtract\\
\bottomrule
\end{tabular}
\end{table}

\begin{table}[ht]
\small
\centering
\caption{Types of Operands}
\label{table: Types of operands}
\begin{tabular}{cccc}
\toprule
\textbf{Type} & \textbf{Operand 1} & \textbf{Operand 2} & \textbf{\# of Operands} \\
\midrule
none & - & - & 0 \\ \hline
addr & address & - & 1 \\ \hline
ref & \begin{tabular}{@{}l@{}}memory\\ reference\end{tabular} & - & 1 \\ \hline
reg-reg & register & register & 2 \\ \hline
reg-addr & register & register & 2 \\\hline
reg-cnst & register & \begin{tabular}{@{}l@{}}constant \\ value\end{tabular} & 2 \\\hline
reg-ref & register & \begin{tabular}{@{}l@{}}memory \\ reference\end{tabular} & 2 \\\hline
ref-cnst & \begin{tabular}{@{}l@{}}memory \\ reference\end{tabular} & \begin{tabular}{@{}l@{}}constant \\ value\end{tabular} & 2 \\\hline
ref-reg & \begin{tabular}{@{}l@{}}memory \\ reference\end{tabular} & register & 2 \\\hline
tri & - & - & 3 \\ \bottomrule
\end{tabular}
\end{table}

\section{More Figures in Evaluations} \label{section:more_figs}

\begin{figure}[ht]
\centering
\includegraphics[width=0.7\linewidth]{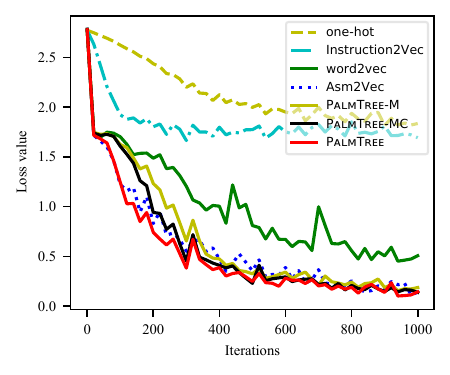}
\vspace{-0.2in}
\caption{Loss value during training}
\label{figure: loss value of Eklavya}
\end{figure}

\begin{figure}[ht]
\centering
\includegraphics[width=0.7\linewidth]{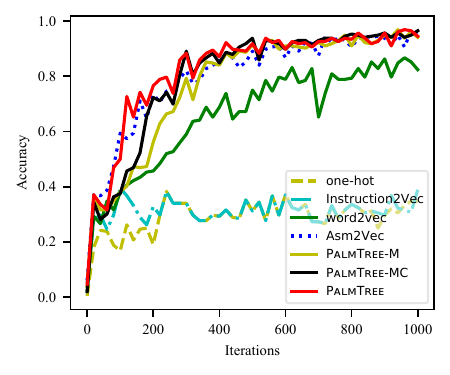}
\vspace{-0.2in}
\caption{Accuracy during training}
\label{figure: accuracy of Eklavya in training}
\end{figure}


\autoref{figure: loss value of Eklavya} and \autoref{figure: accuracy of Eklavya in training} show the results of EKLAVYA in the Function Type Signature Inference task. \autoref{figure: loss value of Eklavya} is the loss value curves of EKLAVYA during training. \autoref{figure: accuracy of Eklavya in training} shows the accuracy curves during the training.


\section{Hyperparameters}

\subsection{Embedding sizes}\label{subsec:embedding_size}

In this experiment, we evaluate the performance of \codename with different embedding sizes. Here we use 64, 128, 256, and 512 as instruction sizes, which is the same as the previous experiment. We test these 4 models on our intrinsic evaluation tasks.

\autoref{table: Embedding sizes} shows all of the results of intrinsic evaluation when having different embedding sizes. From the results, we can observe that there is a clear trend that the performance becomes better when increasing the embedding size. The largest embedding size has the best performance in all three metrics. However, considering efficiency, we recommend having a suitable embedding size configuration according to the hardware capacities. For example, we only have a single GPU (GTX 2080Ti) in our server, thus we chose 128 as the embedding size.

\begin{table}[ht]
\centering
\small
\caption{Embedding sizes}
\label{table: Embedding sizes}
\begin{tabular}{lccccc}
\toprule
& \multicolumn{2}{c}{\textbf{\begin{tabular}[c]{@{}c@{}}opcode outlier \\ detection\end{tabular}}} 
& \multicolumn{2}{c}{\textbf{\begin{tabular}[c]{@{}c@{}}operand outlier \\ detecion\end{tabular}}} 
& \multicolumn{1}{c}{\textbf{\begin{tabular}[c]{@{}c@{}}basicblock \\ sim search\end{tabular}}} \\ \cline{2-6} 
\multirow{-3}{*}{\textbf{\begin{tabular}[c]{@{}l@{}}Embedding\\ Sizes\end{tabular}}} & \multicolumn{1}{c}{Avg.} & Stdev.  & \multicolumn{1}{c}{Avg.}  & Stdev. & \multicolumn{1}{c}{AUC}  \\ \hline

64           &        { 0.836} & {0.0588 } &        { 0.940} & {0.0387 } &        { 0.917} \\
\textbf{128} &        { 0.871} & {0.0440 } &        { 0.944} & {0.0343 } &        { 0.922} \\
256          &        { 0.848} & {0.0560 } &        { 0.954} & {0.0343 } &        { 0.929} \\
512          & \textbf{ 0.878} & {0.0525 } & \textbf{ 0.957} & {0.0335 } & \textbf{ 0.929} \\ 

\bottomrule
\end{tabular}
\end{table}

\subsection{Output layer configurations}\label{subsec:output_layer}
In this experiment, we evaluate the performance of \codename with different output layer configurations. It means that we select a different layer of the transformer model as the output of \codename. By default, \codename uses the second-last layer as the output layer. And we evaluate five different settings, which are the last layer, the second-last layer, the third-last layer, and the fourth-last layer, on our intrinsic evaluation tasks. The embedding size in this experiment is set as 128.

\begin{table}[ht]
\centering
\small
\caption{Output layer configurations}
\label{table: Output layer configurations}
\begin{tabular}{lccccc}
\toprule
\multirow{2}{*}{\textbf{Layers}} & \multicolumn{2}{c}{\begin{tabular}[c]{@{}c@{}}\textbf{opcode outlier} \\ \textbf{detection}\end{tabular}} & \multicolumn{2}{c}{\begin{tabular}[c]{@{}c@{}}\textbf{operand outlier} \\ \textbf{detecion}\end{tabular}} & \begin{tabular}[c]{@{}c@{}}\textbf{basicblock} \\ \textbf{sim search}\end{tabular} \\ \cline{2-6} 
                  & Avg.  & Stdev. & Avg.  & Stdev. & AUC   \\ \midrule
last              &         0.862  & 0.0460 & \textbf{0.982} & 0.0140 &         0.915  \\
\textbf{2nd-last} & \textbf{0.871} & 0.0440 &         0.944  & 0.0343 & \textbf{0.922} \\
3rd-last          &         0.868  & 0.0391 &         0.956  & 0.0287 &         0.918  \\
4th-last          &         0.866  & 0.0395 &         0.961  & 0.0248 &         0.913  \\ 
\bottomrule
\end{tabular}
\end{table}

\autoref{table: Output layer configurations} shows all of the results of the intrinsic metrics when having a different layer as the output layer. There is no obvious advantage to choose any layer as the output layer. However, the second-last layer has the best results in opcode outlier detection and basicblock similarity search. Thus we chose the second-last layer as the output layer in this paper.

\subsection{Context window for CWP}\label{subsec:context_window_size}

\begin{table}[h]
\centering
\small
\caption{Context Window Sizes}
\label{table: Context Window configurations}
\begin{tabular}{lccccccc}
\toprule
& \multicolumn{2}{c}{\textbf{\begin{tabular}[c]{@{}c@{}}opcode \\ outlier \end{tabular}}} & \multicolumn{2}{c}{\textbf{\begin{tabular}[c]{@{}c@{}}operand \\ outlier \end{tabular}}} & \multicolumn{1}{c}{\textbf{\begin{tabular}[c]{@{}c@{}}bb sim \\ search\end{tabular}}} & \multicolumn{2}{c}{\textbf{EKLAVYA}} \\ \cline{2-8} 
\multirow{-3}{*}{\textbf{Sizes}} & \multicolumn{1}{c}{Avg.} & Stdev.  & \multicolumn{1}{c}{Avg.} & Stdev. & \multicolumn{1}{c}{AUC} & \multicolumn{1}{c}{Avg.} & Stdev.  \\ \hline
1          &        {0.864} & {0.0467} & \textbf{0.962} & {0.0168}  & \textbf{0.923} &  {0.930} & {0.0548} \\
\textbf{2} & \textbf{0.871} & {0.0440} &        {0.944} & {0.0343}  &        {0.922} &  \textbf{0.945} & {0.0476} \\
3          &        {0.849} & {0.0444} &        {0.873} & {0.0514}  &        {0.916} &  {0.908} & {0.0633}  \\
4          &        {0.864} & {0.0440} &        {0.957} & {0.0238}  &        {0.914} &  {0.916} & {0.0548}  \\ 
\bottomrule
\end{tabular}
\end{table}

In this experiment, we evaluate the performance of \codename with different context window sizes in the CWP task. For instance, if the context window size is 2, it means that we consider $n-2$, $n-1$, $n+1$ and $n+2$ as contextual instruction when given instruction $n$ as a sample. We evaluate 1, 2, 3, and 4 as four different context window sizes in this experiment.
\autoref{table: Context Window configurations} shows all of the results of the intrinsic metrics when training \codename with different context window configurations. We can observe that context window size 1 and 2 have similar performance on the three intrinsic evaluation metrics, but context window size 2 has the best performance on the downstream task EKLAVYA. Further increasing the context window size to 3 and 4 will lead to worse results. Based on these results, we choose the context window size to be 2.


\end{document}